Data Article

# A dataset for multi-sensor drone detection

Fredrik Svanström[a], Fernando Alonso-Fernandez[b,∗], Cristofer Englund[b,c]

[a] *Air Defence Regiment, Swedish Armed Forces, Sweden*
[b] *Center for Applied Intelligent Systems Research (CAISR), Halmstad University, Halmstad SE 301 18, Sweden*
[c] *RISE, Lindholmspiren 3A, Gothenburg SE 417 56, Sweden*

**A B S T R A C T**

The use of small and remotely controlled unmanned aerial vehicles (UAVs), referred to as drones, has increased dramatically in recent years, both for professional and recreative purposes. This goes in parallel with (intentional or unintentional) misuse episodes, with an evident threat to the safety of people or facilities [1]. As a result, the detection of UAV has also emerged as a research topic [2].

Most of the existing studies on drone detection fail to specify the type of acquisition device, the drone type, the detection range, or the employed dataset. The lack of proper UAV detection studies employing thermal infrared cameras is also acknowledged as an issue, despite its success in detecting other types of targets [2]. Beside, we have not found any previous study that addresses the detection task as a function of distance to the target. Sensor fusion is indicated as an open research issue as well to achieve better detection results in comparison to a single sensor, although research in this direction is scarce too [3–6].

To help in counteracting the mentioned issues and allow fundamental studies with a common public benchmark, we contribute with an annotated multi-sensor database for drone detection that includes infrared and visible videos and audio files. The database includes three different drones, a small-sized model (Hubsan H107D+), a medium-sized drone (DJI Flame Wheel in quadcopter configuration), and a performance-grade model (DJI Phantom 4 Pro). It also includes other flying objects that can be mistakenly detected as drones, such as birds, airplanes or helicopters. In addition to using several different sensors, the number of classes is higher than in previous studies [4]. The video part contains 650 infrared and visible videos (365 IR and 285 visible) of drones, birds, airplanes and helicopters. Each clip is of ten seconds, resulting in a total of 203,328 annotated frames. The database is complemented with 90 audio files of the classes drones, helicopters and background noise.

∗ Corresponding author.
*E-mail address:* feralo@hh.se (F. Alonso-Fernandez).

To allow studies as a function of the sensor-to-target distance, the dataset is divided into three categories (Close, Medium, Distant) according to the industry-standard Detect, Recognize and Identify (DRI) requirements [7], built on the Johnson criteria [8]. Given that the drones must be flown within visual range due to regulations, the largest sensor-to-target distance for a drone in the dataset is 200 m, and acquisitions are made in daylight. The data has been obtained at three airports in Sweden: Halmstad Airport (IATA code: HAD/ICAO code: ESMT), Gothenburg City Airport (GSE/ESGP) and Malmö Airport (MMX/ESMS). The acquisition sensors are mounted on a pan-tilt platform that steers the cameras to the objects of interest. All sensors and the platform are controlled with a standard laptop vis a USB hub.

**Specifications Table**

| | |
|---|---|
| Subject | Computer Science |
| | Computer Vision and Pattern Recognition |
| Specific subject area | The dataset can be used for multi-sensor drone detection and tracking. It contains infrared and visible videos and audio files of drones, birds, airplanes, helicopters, and background sounds. |
| Type of data | Audio (.wav) |
| | Video (.mp4) |
| | Data labels in Matlab files (.mat) |
| | Excel file |
| How data were acquired | Three different drones are used to collect and compose the dataset: Hubsan H107D+, a small-sized first-person-view (FPV) drone, the high-performance DJI Phantom 4 Pro, and finally, the medium-sized kit drone DJI Flame Wheel in quadcopter (F450) configuration. |
| | The thermal infrared camera used (IRcam) is a FLIR Breach PTQ-136 using the FLIR Boson 320 × 256 pixels sensor (Y16 with 16-bit greyscale). The field of view (FoV) is 24° horizontally and 19° vertically. The output is sent via a USB-C port at a rate of 60 frames per second (FPS). |
| | A Sony HDR-CX405 video camera (Vcam) is used to record the scenes in the visible range. The output is an HDMI signal, and hence a frame grabber in the form of an Elgato Cam Link 4K is used to capture a 1280 × 720 video stream in YUY2- format (16 bits per pixel) at 50 FPS. The Vcam is set to have about the same field of view as the IRcam. |
| | Due to the limited FoV of the infrared and visible cameras, they are steered by an ELP 8 megapixel fish-eye lens camera (Fcam) covering 180° horizontally and 90° vertically. This outputs a 1024 × 768 video stream in Mjpg-format at 30 FPS via a USB connector. |
| | To capture the sound of the distinct classes of the database, a Boya BY-MM1 mini cardioid directional microphone is employed. |
| | The sensors are mounted on a pan/tilt platform Servocity DDT-560H direct drive. Pan/tilt motion is achieved with two Hitec HS-7955TG servos. A Pololu Mini Maestro 12-Channel USB servo controller is included. The computational part is done on a Dell Latitude 5401 laptop equipped with an Intel i7-9850H CPU and an Nvidia MX150 GPU. The laptop is connected to all the sensors mentioned above and the servo controller using the built-in ports and an additional USB hub. |
| Data format | Raw |
| Parameters for data collection | The videos are recorded at locations in and around Halmstad and Falkenberg (Sweden), at Halmstad Airport (IATA code: HAD/ICAO code: ESMT), Gothenburg City Airport (GSE/ESGP) and Malmö Airport (MMX/ESMS). |
| | Since the drones must be flown within visual range due to regulations, the data is recorded in daylight. For the same reason, the maximum distance between the drones and the sensors is 200 m. |
| Description of data collection | The fish-eye lens camera is used to feed a foreground/background detector that produces binary masks of moving objects. A multi-object Kalman filter tracker then steers the infrared and visible cameras via a servo controller mounted on a pan/tilt platform. |
| | To get a more comprehensive dataset, a small portion of the database is from non-copyrighted material from the YouTube channel "Virtual Airfield operated by SK678387". This is because it has not been possible to film all types of targets since this work has been done during the drastic reduction of operations due to the COVID pandemic. |
| Data source location | Institution: School of Information Technology, Halmstad University<br>City/Town/Region: Halmstad<br>Country: Sweden<br>Latitude and longitude (and GPS coordinates, if possible) for collected samples/data: 56° 39′ 30.59″ N, 12° 52′ 25.79″ E |
| Data accessibility | Repository name: Zenodo<br>Data identification number: [provide number, if available]<br>Direct URL to data: http://dx.doi.org/10.5281/zenodo.5500576 |
| Related research article | F. Svanström, C. Englund, F. Alonso-Fernandez, "Real-Time Drone Detection and Tracking with Visible, Thermal and Acoustic Sensors", *2020 25th International Conference on Pattern Recognition (ICPR)*, Milan, Italy, 10–15 January 2021, pp. 7265–7272. http://dx.doi.org/10.1109/ICPR48806.2021.9413241 |

**Value of the Data**

- The dataset can be used to develop new algorithms for drone detection using multi-sensor fusion from infrared and visible videos and audio files.
- The dataset can be used by scientists in signal/image processing, computer vision, artificial intelligence, pattern recognition, machine learning and deep learning fields.
- The provided data can help in developing systems that distinguish drones from other objects that can be mistaken for a drone, such as birds, airplanes or helicopters.
- This dataset can be used to build a drone detection system, which can aid in preventing threatening situations where the security of people or facilities can be compromised, such as flying over restricted areas in airports or crowds in cities.

## 1. Data Description

The database includes data captured with a thermal infrared camera (IRcam), a camera in the visible range (Vcam), and a microphone. The classes available with each type of sensor are indicated in Table 1. The sensors employed are specified in the next section. Both the videos and the audio files are cut into ten-second clips to be easier to annotate.

**Table 1**
Classes of the different sensors.

| Sensor | Output classes | | | |
|---|---|---|---|---|
| IRcam | Airplane | Bird | Drone | Helicopter |
| Vcam | Airplane | Bird | Drone | Helicopter |
| Audio | | | Drone | Helicopter | Background |

## 1.1. Video dataset

Overall, the video dataset contains 650 videos (365 IR and 285 visible, of ten seconds each), with a total of 203328 annotated frames. The IR videos have a resolution of 320 × 256 pixels, whereas the visible videos have 640 × 512. The largest distance between the sensors and a drone in the database is 200 m. All videos are in mp4 format. The filenames start with the sensor type, followed by the target type and a number, e.g. IR_DRONE_001.mp4. The annotation of the respective clips has the additional tag LABELS, e.g. IR_DRONE_001_LABELS.mat. These files are Matlab Ground-Truth objects and using the Matlab video labeller app, the videos and respective label files can easily be opened, inspected, and even edited. If the dataset is to be used in another development environment, the label files can be opened in Matlab, and the content is saved in the desired format, such as .csv. Importing Matlab files into a Python environment can also be done using the scipy.io.loadmat command.

## 1.2. Audio dataset

The audio part has 90 ten-second files in wav-format with a sampling frequency of 44100 Hz. There are 30 files of each of the three output audio classes indicated in Table 1. The clips are annotated with the filenames themselves, e.g. DRONE_001.wav, HELICOPTER_001.wav, BACKGROUND_001.wav, etc. The audio in the dataset is taken from the videos or recorded separately. The background sound class contains general background sounds recorded outdoor in the acquisition location and includes some clips of the sounds from the servos moving the pan/tilt platform where the sensors were mounted.

## 1.3. Distance categories in the video dataset

The distribution of videos among the four output video classes is shown in Tables 2 and 3. Since one of the objectives of this work is to explore performance as a function of the sensor-to-target distance, the video dataset has been divided into three distance category bins: Close, Medium and Distant. The borders between these bins are chosen to follow the industry-standard Detect, Recognize and Identify (DRI) requirements [7], building on the Johnson criteria [8], as shown in Fig. 1. Since the distance bin information of the clip is not included in the filename, there is also an associated excel-sheet where this is shown in a table. This table also contains information about the exact drone type and if the clip comes from the Internet or not.

**Table 2**
The distribution of the 365 IR videos.

| | Class | | | |
|---|---|---|---|---|
| Bin | AIRPLANE | BIRD | DRONE | HELICOPTER |
| Close | 9 | 10 | 24 | 15 |
| Medium | 25 | 23 | 94 | 20 |
| Distant | 40 | 46 | 39 | 20 |

**Table 3**
The distribution of the 285 visible videos.

| | Class | | | |
|---|---|---|---|---|
| Bin | AIRPLANE | BIRD | DRONE | HELICOPTER |
| Close | 17 | 10 | 21 | 27 |
| Medium | 17 | 21 | 68 | 24 |
| Distant | 25 | 20 | 25 | 10 |

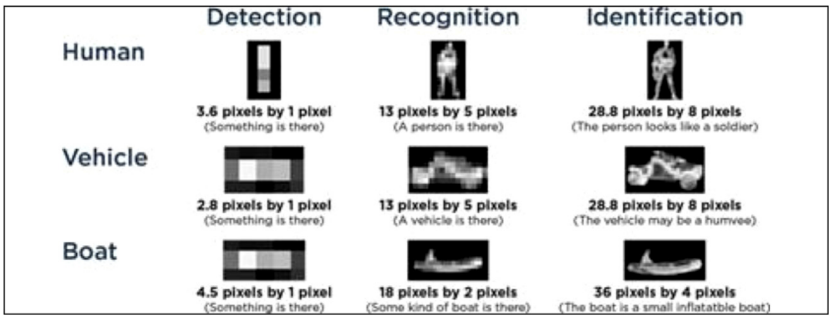

**Fig. 1.** The DRI-requirements. From [7].

**Table 4**
The distance bin division for the different target classes.

| | Class | | | |
|---|---|---|---|---|
| Bin | AIRPLANE | BIRD | DRONE | HELICOPTER |
| Close | <1000 m | <40 m | <20 m | <500 m |
| Medium | 1000–3000 m | 40–120 m | 20–60 m | 500–1500 m |
| Distant | >3000 m | >120 m | >60 m | >1500 m |

The Close distance bin is from 0 m out to a distance where the target is 15 pixels wide in the IRcam image, i.e. the requirement for recognition according to DRI. The Medium bin stretches from where the target is from 15 down to 5 pixels, hence around the DRI detection point, and the Distant bin is beyond that. Given the resolution and field of view of the IRcam and the object class sizes: Drone 0.4 m, bird 0.8 m, helicopter[1] 10 m and airplane[2] 20 m, we get a distance division for the different object types summarized in Table 4.

To illustrate the *detect, recognize, and identify* concept, objects from all the target classes being 15 pixels in width are shown in Fig. 2. At this level, we can not only detect but also recognize the different objects, albeit without necessarily identifying them, i.e. explicitly telling what kind of helicopter it is and so on.

## 2. Experimental Design, Materials and Methods

### 2.1. Dataset composition

To compose the dataset, three different drones are used. These are of the following types: Hubsan H107D+, a small first-person-view (FPV) drone; the high-performance DJI Phantom 4

---
[1] Bell 429, one of the helicopter types in the dataset, has a length of 12.7 m.
[2] Saab 340 has a length of 19.7 m and a wingspan of 21.4 m.

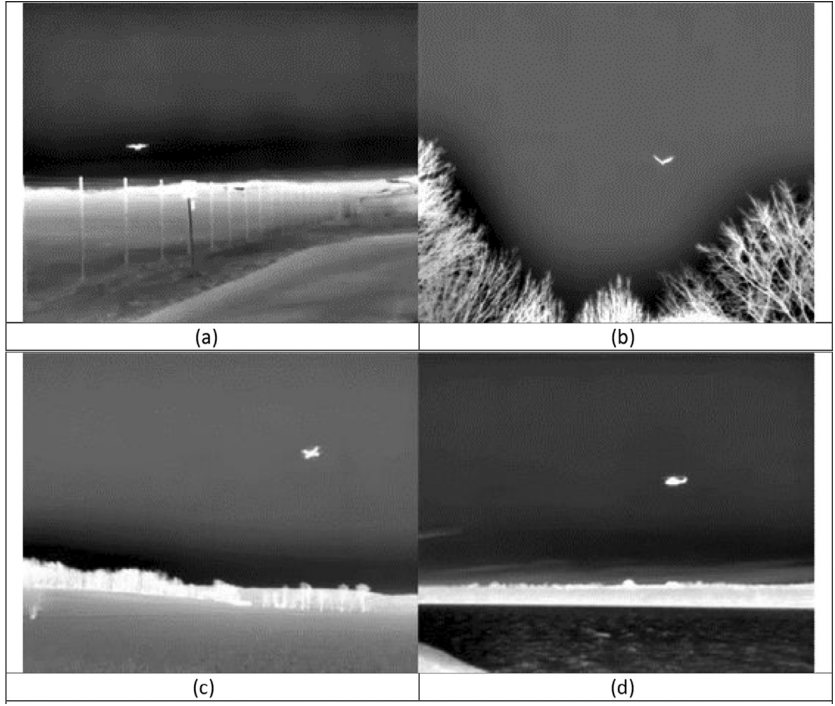

**Fig. 2.** Objects on the limit between the close and medium distance bins. (a) An airplane at a distance of 1000 m. (b) A bird at a distance of 40 m. (c) A drone at at distance of 20 m. (d) A helicopter at a distance of 500 m.

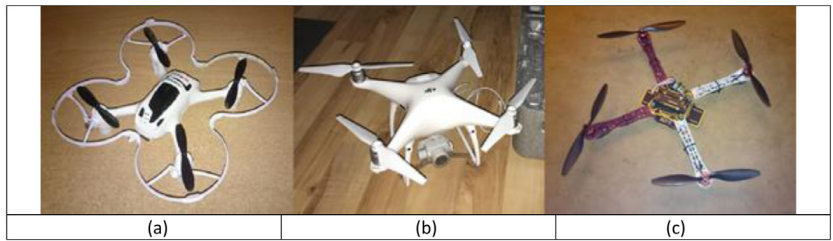

**Fig. 3.** The three drone types of the dataset. (a) Hubsan H107D+. (b) DJI Phantom 4 Pro. (c) DJI Flame Wheel F450.

Pro; and the medium-sized DJI Flame Wheel. The latter can be built both as a quadcopter (F450) or in a hexacopter configuration (F550). The version used in this work is an F450 quadcopter. All three types can be seen in Fig. 3. These drones differ in size, with Hubsan H107D+ being the smallest, with a side length from motor-to-motor of 0.1 m. The Phantom 4 Pro and the DJI Flame Wheel F450 are slightly larger with 0.3 and 0.4 m motor-to-motor side lengths, respectively.

Since the drones must be flown within visual range, the largest sensor-to-target distance for a drone is 200 m. There are also eight clips (five IR and three visible videos) within the dataset with two drones flying simultaneously, as shown, for example, in Fig. 4.

Common birds appearing in the dataset are the rook (*Corvus frugilegus*) and the western jackdaw (*Coloeus monedula*) of the crow family (*Corvidae*), the European herring gull (*Larus argentatus*), the common gull (*Larus canus*) and the black-headed gull (*Chroicocephalus ridibundus*) of the *Laridae* family of seabirds. Occasionally occurring in the dataset are also the black kite (*Milvus migrans*) of the *Accipitridae* family and the Eurasian skylark (*Alauda arvensis*) of the lark family

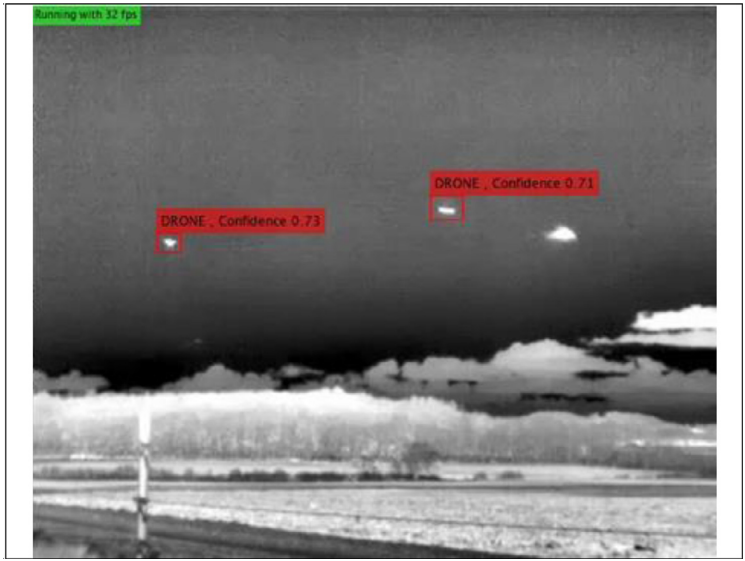

**Fig. 4.** Example of IR video with two drones appearing in the image.

(*Alaudidae*). On average, these species have a wingspan of 0.8 m, making them about twice the size of the medium-sized consumer grade drone.

To get a more comprehensive dataset, both in terms of aircraft types and sensor-to-target distances, our data has been completed with non-copyrighted material from the YouTube channel "Virtual Airfield operated by SK678387" [9], in particular 11 plus 38 video clips in the airplane and helicopter categories, respectively. This is because it has not been possible to film all types of suitable targets, given that this work has been carried out during the drastic reduction of flight operations due to the COVID19 pandemic.

The drones and helicopters appearing in the database move in most cases at normal flying speeds (in the range of 0–60 km/h for drones, and 0–300 km/h for helicopters). In some few cases, these vehicles fly at very low speed or are hovering.

## 2.2. Location and data collection

The captured videos are recorded at locations in and around Halmstad and Falkenberg (Sweden), at Halmstad Airport (IATA code: HAD/ICAO code: ESMT), Gothenburg City Airport (GSE/ESGP) and Malmö Airport (MMX/ESMS).

The drone flights are all done in compliance with the Swedish national rules for unmanned aircraft found in [10]. The most critical points applicable to the drones and locations used in this work are:

- √ When flown, the unmanned aircraft shall be within its operational range and well within the pilot's visual line of sight.
- √ When flown in uncontrolled airspace, the drone must stay below 120 m from the ground.
- √ When flying within airports' control zones or traffic information zones and if you do not fly closer than 5 km from any section of the airport's runway(s), you may fly without clearance if you stay below 50 m from the ground.
- √ For the protection of people, animals and property which are unrelated to the flight, there must be a horizontal safety distance between these and the unmanned aircraft throughout the flight.

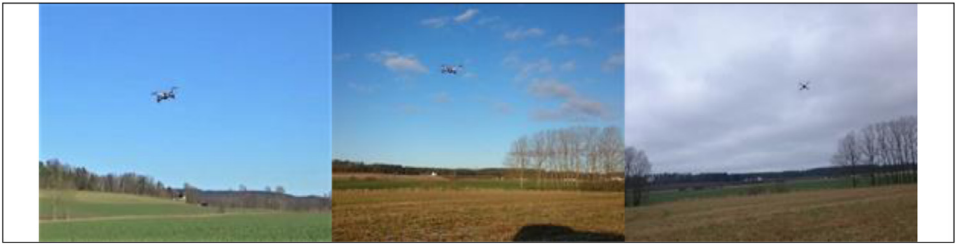

**Fig. 5.** Examples of varying weather conditions in the dataset.

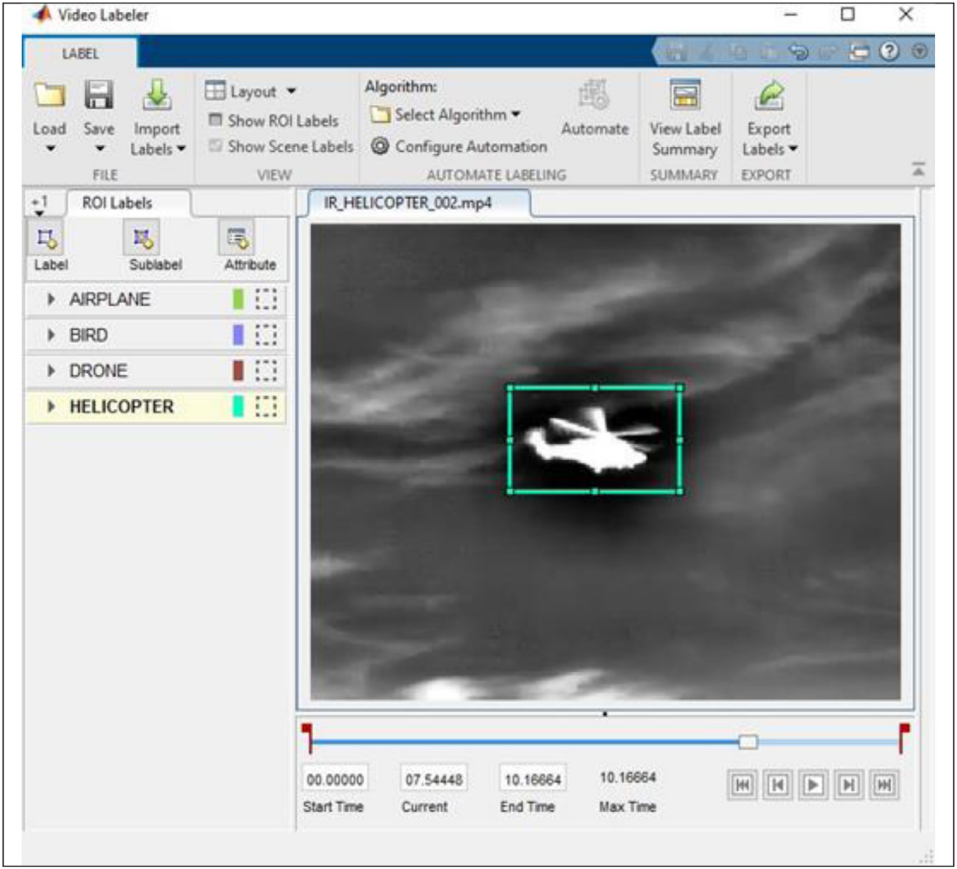

**Fig. 6.** The Matlab video labeller app.

Since the drones must be flown within visual range, the dataset is recorded in daylight, even if the thermal and acoustic sensors could be used even in complete darkness. The weather in the dataset stretches from clear and sunny to scattered clouds and completely overcast, as shown in Fig. 5.

The annotation of the video dataset is done using the Matlab video labeller app. An example from a labeling session is shown in Fig. 6.

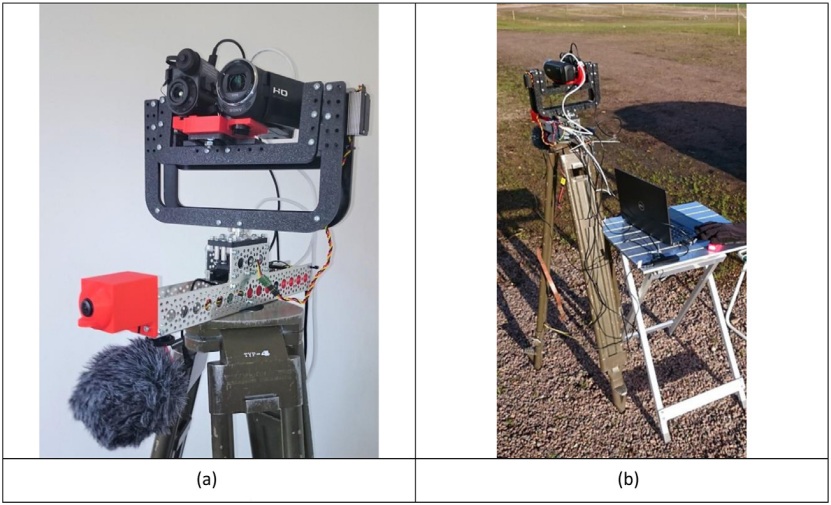

**Fig. 7.** Setup of the acquisition system. (a) The main parts of the system. On the lower left is the microphone, and above that is the fish-eye lens camera. On the pan and tilt platform in the middle are the IR- and video cameras. The holder for the servo controller and power relay boards is placed behind the pan servo inside the aluminium mounting channel. (b) The system deployed just north of the runway at Halmstad airport (IATA/ICAO code: HAD/ESMT).

## 2.3. Acquisition hardware

The captured data is from a thermal infrared camera (IRcam), a camera in the visible range (Vcam), and a microphone. They are placed together on a pan/tilt platform that can be aimed in specific directions. Due to the limited field of view of these cameras, they are steered towards specific directions guided by a fish-eye lens camera (Fcam) covering 180° horizontally and 90° vertically. The role of the fish-eye camera is not to detect specific classes but to detect moving objects in its field of view. If nothing is detected by the Fcam, the platform can be set to move in two different search patterns to scan the sky around the system.

The Fcam is used to feed a foreground/background detector based on Gaussian Mixture Models (GMM), which produces binary masks of moving objects. This is followed by a multi-object Kalman filter tracker, which, after calculating the position of the best-tracked target, sends the azimuth and elevation angles. The best-tracked target is defined as the one with the longest track history. Based on this, the pan/tilt platform servos are then steered via the servo controller so that the moving object can be captured by the infrared and visible cameras.

All computations are made on a standard laptop. Fig. 7 shows the main parts of the system. To have a stable base, all hardware components, except the laptop, are mounted on a standard surveyor's tripod. This also facilitates transport and deployment outdoors, as shown in the right part of the figure.

The thermal infrared camera (IRcam) employed is a FLIR Breach PTQ-136 with the FLIR Boson sensor, having 320 × 256 pixels of resolution. The field of view of the IR-camera is 24° horizontally and 19° vertically. Fig. 8 shows an image taken from the IRcam video stream. The IRcam has two output formats, a raw 320 × 256 pixels format (Y16 with 16-bit greyscale) and an interpolated 640 × 512 pixels image in the I420 format (12 bits per pixel). The color palette can be changed for the interpolated image format, and several other image processing features are also available. The raw format is used in the database to avoid the extra overlaid text information of the interpolated image. The output from the IRcam is sent to the laptop via a USB-C port at a rate of 60 frames per second (FPS). The IRcam is also powered via a USB connection. Notably, the Boson sensor of the FLIR Breach has a higher resolution than the one used in [11] where a FLIR Lepton sensor with 80 × 60 pixels was used. In that paper, the authors were able to detect three different drone types up to 100m. However, this detection was done manually by a person looking at the live video stream.

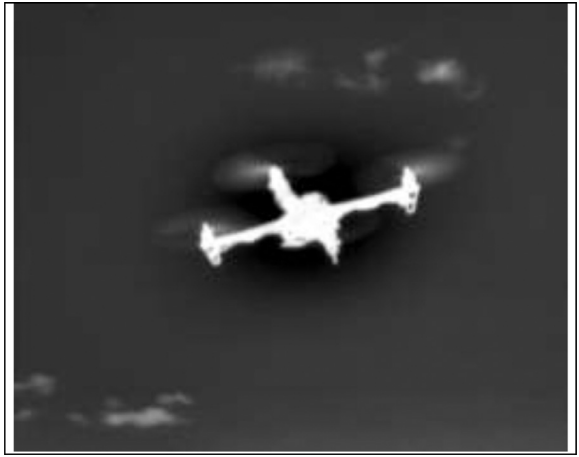

**Fig. 8.** The F450 drone flying at a distance of 3 m in front on the IRcam.

To record data in the visible range of the spectrum, a Sony HDR-CX405 video camera (Vcam) is used, which provides data through an HDMI port. To feed the laptop via USB, we use an Elgato Cam Link 4K frame grabber, which gives a 1280 × 720 video stream in YUY2- format (16 bits per pixel) at 50 FPS. Due to its adjustable zoom lens, the field of view of the Vcam can be set to different values, which in this work is set to about the same field of view as the IRcam.

To monitor a larger part of the surroundings of the system, an ELP 8 megapixel 180° fish-eye lens camera (Fcam) is also used. This outputs a 1024 × 768 video stream in Mjpg-format at 30 FPS via a USB connector.

The sound of the distinct classes of the database is captured with a Boya BY-MM1 mini cardioid directional microphone, which is also connected to the laptop.

As mentioned, the IR- and video cameras are mounted on a pan/tilt platform. This is the Servocity DDT-560H direct drive tilt platform together with the DDP-125 Pan assembly, also from Servocity. To achieve the pan/tilt motion, two Hitec HS-7955TG servos are used. A Pololu Mini Maestro 12-Channel USB servo controller is included so that the respective position of the servos can be controlled from the laptop. Since the servos have shown a tendency to vibrate when holding the platform in specific directions, a third channel of the servo controller is also used to give the possibility to switch on and off the power to the servos using a small optoisolated relay board. To supply the servos with the necessary voltage and power, both a net adapter and a DC-DC converter are available. The DC-DC solution is used when the system is deployed outdoors and, for simplicity, it uses the same battery type as one of the available drones. Some other parts from Actobotics are also used in the mounting of the system, and the following has been designed and 3D-printed: adapters for the IR-, video- and fish-eye lens cameras, and a case for the servo controller and power relay boards.

All computations and acquisitions are made on a Dell Latitude 5401 laptop, having an Intel i7-9850H CPU and an Nvidia MX150 GPU. All the sensors mentioned above and the servo controller are connected to the laptop using the built-in ports and an additional USB hub.

### 2.4. Dataset use

The intended purpose of the dataset is to be used in the training and evaluation of stationary drone detection systems on the ground. It should be possible, however, to use the database (or parts of it) on-board a drone if, for example, the purpose of such drone is to find other drones.

The drone detection system used in this project utilized several sensors at the same time, including sensor fusion. Therefore, the computational cost is relatively high, and hence a laptop with a separate GPU was used. It might be possible to use a simple microcontroller if the drone detection system trained and evaluated with the dataset uses only one sensor or a small number of them. If the detection system is to be placed, for example, on-board a drone, it must also be considered that it would affect battery duration, reducing the effective flying time of the drone.

The data contained in the database can be used as-is without filtering or enhancement.

## Ethics Statement

The data does not include human subjects or animals.

## Declaration of Competing Interest

The authors declare that they have no known competing financial interests or personal relationships which have or could be perceived to have influenced the work reported in this article.

## CRediT Author Statement

**Fredrik Svanström:** Conceptualization, Methodology, Investigation, Data curation, Writing – review & editing; **Fernando Alonso-Fernandez:** Conceptualization, Supervision, Funding acquisition, Writing – original draft; **Cristofer Englund:** Conceptualization, Supervision, Writing – review & editing.

## Acknowledgments

This work has been carried out by Fredrik Svanström in the context of his Master Thesis at Halmstad University (Master's Programme in Embedded and Intelligent Systems). Author F. A.-F. thanks the Swedish Research Council and VINNOVA for funding his research.